\documentclass[letterpaper, 10 pt, conference]{ieeeconf}  % Comment this line out if you need a4paper

\IEEEoverridecommandlockouts                              % This command is only needed if 
\usepackage{graphicx} % for pdf, bitmapped graphics files
\usepackage{amsmath} % assumes amsmath package installed
\usepackage{amssymb}  % assumes amsmath package installed
\usepackage{subcaption}
\usepackage{float}
\usepackage{booktabs}
\usepackage{siunitx}
\usepackage[utf8]{inputenc}
\usepackage[T1]{fontenc}

\usepackage{bm}

\usepackage{tikz}

\usepackage{lipsum}

\usepackage{xparse}

\NewDocumentCommand\bbm{}{ \begin{bmatrix} }
\NewDocumentCommand\ebm{}{ \end{bmatrix} }

\NewDocumentCommand\Matrix{m}{ \boldsymbol{\mathbf{#1}} }

\NewDocumentCommand\Real{}{ \mathbb{R} }

\NewDocumentCommand\LieGroupSE{m}{ \mathrm{SE}(#1) }
\NewDocumentCommand\LieAlgebraSE{m}{ \mathfrak{se}(#1) }

\NewDocumentCommand\CoordinateFrame{m}{ \underrightarrow{\Matrix{\mathcal{F}}}_{#1} }

\NewDocumentCommand\Transform{}{ \Matrix{T} }

\usepackage{balance}

\title{\LARGE \bf
    RaSCL: Radar to Satellite Crossview Localization
}

\author{Blerim Abdullai$^{1}$, Tony Wang, Aoran Jiao, Xinyuan Qiao, Florian Shkurti, and Timothy D. Barfoot
\thanks{The authors are with the University of Toronto Robotics Institute}
 \thanks{$^{1}$ {Main Contact: \tt  blerim@cs.toronto.edu}}%
}

\begin{document}
\maketitle
\thispagestyle{empty}
\pagestyle{empty}

%%%%%%%%%%%%%%%%%%%%%%%%%%%%%%%%%%%%%%%%%%%%%%%%%%%%%%%%%%%%%%%%%%%%%%%%%%%%%%%%
\begin{abstract}

    GNSS is unreliable, inaccurate, and insufficient in many real-time autonomous field applications. In this work, we present a GNSS-free global localization solution that contains a method of registering imaging radar on the ground with overhead RGB imagery, with joint optimization of relative poses from odometry and global poses from our overhead registration. Previous works have used various combinations of ground sensors and overhead imagery, and different feature extraction and matching methods. These include various handcrafted and deep-learning-based methods for extracting features from overhead imagery. Our work presents insights on extracting essential features from RGB overhead images for effective global localization against overhead imagery using only ground radar and a single georeferenced initial guess. We motivate our method by evaluating it on datasets in diverse geographic conditions and robotic platforms, including on an Unmanned Surface Vessel (USV) as well as urban and suburban driving datasets.
    %\sam{I think it is nessssary to include some of the results such as we achieved blah blah result on our own boat sequences and also other sequences.}

    %\sam{Also do you plan to open source the code, you can include a link here and claim as one of the contributions}
%\textbf{Note: We need to rewrite the abstract}
\end{abstract}

%%%%%%%%%%%%%%%%%%%%%%%%%%%%%%%%%%%%%%%%%%%%%%%%%%%%%%%%%%%%%%%%%%%%%%%%%%%%%%%%

\section{INTRODUCTION}

In GPS-denied or degraded environments, effective localization proves to be difficult. Real-time GPS measurements are susceptible to atmospheric conditions, satellite positions, and multipath reflections. Real-time Kinematic (RTK) GPS can mitigate these issues. However, this requires line-of-sight communication or a cellular connection, which is often infeasible in many remote field robotic deployments. A common approach is to localize against a prior map \cite{BurnettAreWeReady2022}. However, these maps require a prior traversal, limiting localization to only pre-mapped areas. Simultaneous Localization and Mapping (SLAM) seeks to solve this by optimizing for both ego-centric poses and the locations of landmarks in a global frame. Yet, these methods scale poorly with the size of the environment and can accumulate errors over time. 

Instead, we propose localizing against RGB satellite imagery using a millimeter-wave (mmWave) imaging radar on the ground.
Radio Detection And Ranging (radar) sensors are well-developed sensing systems that use radio waves to detect surrounding environments by measuring their distance/angle, and velocity \cite{skolnik1962introduction}. Thanks to their longer range, radar sensors have been proven suitable for detection in challenging scenarios where camera and lidar systems struggle due to a lack of geometric features and environmental artifacts such as dust, fog, and snow. Their lower wavelength allows them to penetrate through those particles and remain robust to extreme weather conditions \cite{s21165397}. Because of these sensing qualities, the robotics community has investigated radar as an alternative sensor to lidar for perception, object tracking, and localization \cite{qiao2024radar}. 

RGB satellite imagery is widely available, globally consistent, memory-efficient, and does not require a prior traversal of a desired route. Previous work has studied localization against public data using different combinations of ground sensors and overhead imagery representations and is often referred to as \textit{crossview localization}. This can include ground cameras to RGB satellite mages \cite{Fervers_2023_CVPR}, lidar to RGB images \cite{Tang2023} \cite{fervers2022continuous}, sonar to overhead images \cite{mcconnell2022overheadimagefactorsunderwater}, and radar to overhead imagery \cite{tang2020rslnet} \cite{tang2021SelfSuper} \cite{Tang2023} \cite{Hong2023RadarOSM}. Many of the previous works in radar to overhead image localization either only consider single frame evaluation, or use explicit road segment information to localize. 
%However, to the best of our knowledge, a stand-alone learning-based localization system with radar as the only sensor and an initial guess for reliable localization against RGB satellite images for the duration of a survey has not been developed or evaluated before.

\begin{figure}[t]
    \centering
    \begin{minipage}{\linewidth}
        \centering
    \includegraphics[width=\linewidth]{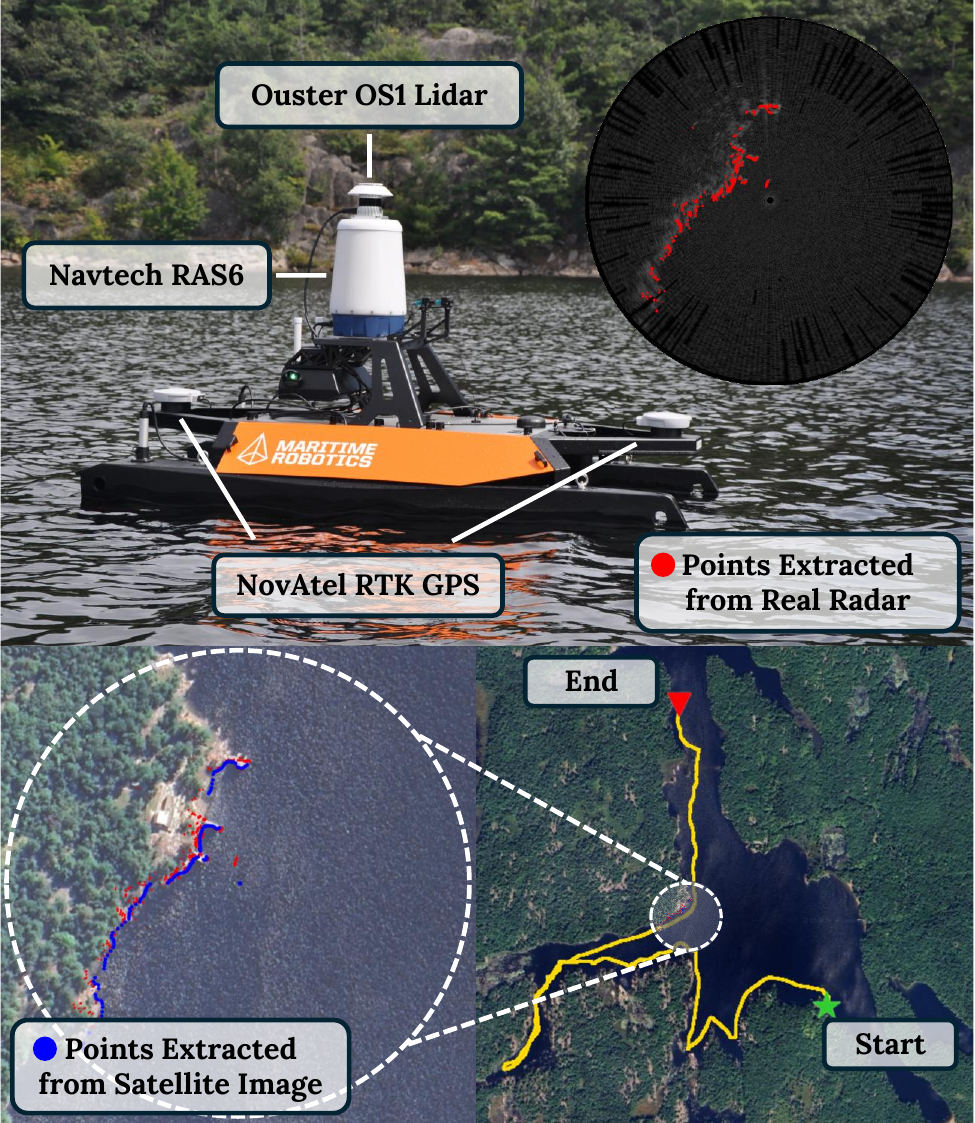}
        \caption{Top shows our Otter USV data collection platform equipped with a Navtech RAS6 radar, an Ouster OS1 lidar and a NovAtel RTK GPS system. Top right illustrates a radar observation and extracted real radar points in red. Bottom left shows the extracted real radar points and our predicted blue points overlaid onto the satellite image against which we are localizing. Bottom right shows the route driven in yellow.}
        \label{fig:moneyshot}
    \end{minipage}
    \vspace*{-5mm}
\end{figure}

\subsection{Contributions}
Our main contributions can be summarized as follows:
\begin{enumerate}
\item A method of registration for a single ground radar
scan to a proximal overhead image using deep learning and Iterative Closest Point (ICP)
without explicit use of lane lines and road segments.
\item Joint optimization of radar odometry and overhead
image registration capable of GNSS-free global localization over unseen terrain using only public imagery, radar, and a georeferenced initial guess.
\item Evaluation on diverse geographic conditions
and robotic platforms, including urban, suburban, and
marine settings.

\end{enumerate}

\section{RELATED WORK}

There exists an abundance of radar sensors, each catering to specific applications. Doppler radar measures the radial velocity of objects in the scene and is commonly used in weather forecasting and speed detection \cite{doviak2014doppler}. Synthetic Aperture Radar (SAR) produces high-resolution images for reconnaissance and mapping \cite{fitch2012synthetic}. Ground-Penetrating Radar (GPR) detects objects underground using high-frequency radio waves \cite{jol2008ground}. Phased-Array Radar uses multiple antennas to electronically steer the beam without moving the antenna \cite{4250281}. In military applications, the Inverse Synthetic Aperture Radar (ISAR) is used to image moving targets such as ships and aircraft \cite{9513303}. Marine radars often use a 5 GHz frequency band for near-port navigation in dense fog and longer ranges \cite{Jang_2024}. Due to the wide variety in frequencies, beam forming, coding functions, and hardware processing techniques, methods developed for one type of radar sensor will often not be suitable for another.  

\subsection{Radar Odometry}
As with most sensors, there are multiple possible methods of radar-to-radar image registration. These include filtering the raw image to obtain a sparse set of points and using ICP \cite{Besl92ICP}, learning-based methods for sparse feature matching, dense methods that compute cross correlation on the raw image for image registration, and hybrid methods that combine sparse and dense registration. Cen et al. \cite{Cen2019Ego} sparked the interest in mmWave radar for robotics and proposed a sparse filtering method to extract a point cloud and a gradient-based graph matching registration method. CFEAR \cite{adolfsson2022LiDAR} is an example of a sparse point cloud pipeline and the current state of the art in radar-only odometry. CFEAR considers the k-strongest return, motion distortion, and refining filtering techniques to significantly improve radar odometry by reducing drift. Barnes et al. \cite{barnes2020radarlearningpredictrobust} proposed a learning-based keypoint detection architecture that learns keypoint locations, descriptors, and a scoremap. The method is supervised using ground truth poses by performing differentiable feature matching to estimate a relative pose. Burnett et al. \cite{burnett2021radarodometrycombiningprobabilistic} proposed a self-supervised pipeline using this architecture. Barnes et al. \cite{barnes2020maskingmovinglearningdistractionfree} use a CNN to mask distractor portions of the radar scan by supervising a correlation-based registration algorithm with ground truth poses. It is also possible to use 5 GHz marine radar for radar odometry as explored in \cite{Jang_2024} using a hybrid approach.  The authors developed a handcrafted descriptor to solve for rotation in a dense fashion, and used filtered contour detection with point normal matching to solve for translation. Schiller et al. \cite{schiller2022} mmWave radar for radar odometry in a marine setting, conducting brute force matching with FAST descriptors. If Doppler velocity information is available from the FMCW radar, odometry performance can be further improved by leveraging the Doppler radial velocity information and the 2-D point cloud extracted from the radar scan images \cite{lisus2024doppler}. 

%Doppler-enabled radar odometry can even operate in a tunnel and skyway where a lack of useful features are present in the radar scans. In learning radar-inertial odometry, milliEgo, a deep fusion architecture that combines a single-chip mmWave FMCW radar with an IMU, was trained by Lu et al \cite{lu2020milliego}. They designed a specialized neural network to regress the 6-DoF pose change from radar data, and introduced a mixed-attention module to robustly fuse radar and inertial measuements. This learning-based radar/IMU odometry achieved~1.3\% drift in trajectory error over distance, generalizing well to unseen environments. The approach was efficient enough for real-time use on embedded platforms. These results underscore that even low-cost radars can yield accurate ego-motion estimates when paired with deep learning and sensor fusion.

\subsection{Radar Localization and Mapping}
In the realm of simultaneous localization and mapping (SLAM), Hong et al. \cite{hong2021radar} introduced a radar-based graph-SLAM system designed for robust performance in large-scale environments and adverse weather conditions. Their approach converts FMCW  radar data into 3D point clouds and employs keyframe-based tracking for real-time pose estimation. For local mapping, the system utilizes bundle adjustment to enhance spatial consistency. Loop closures are detected using a rotation-invariant global descriptor tailored for 3D point cloud recognition. To explore the use of Doppler information in radar localization, Sie et al. \cite{sie2024radarize} propose Radarize, a SLAM pipeline that uses only a single-chip FMCW radar and leverages Doppler motion sensing with learning-based odometry. Radarize employs two ResNet-based CNN models to estimate translation and rotation directly from radar spectral data (Doppler-azimuth and range-azimuth heatmaps). This radar-native approach also includes techniques to suppress multipath ghost reflections, addressing common radar artifacts. Barnes et al. \cite{barnes2018driven} showed that adding radar depth cues to a visual SLAM can help mask out unreliable regions in images. %Should Talk about Keenan Are We Ready Paper.

%\subsection{Public Radar Datasets}
%Several radar-focused datasets have been collected and open-sourced to facilitate research in the domain of radar odometry, radar localization, and radar place recognition. 
%The Boreas dataset \cite{burnett2023boreas}, collected year-round by repeatedly traversing the same urban-rural city route, captures seasonal variations and adverse weather (e.g., rain, snow) across 350+ km of driving data.  The Oxford Radar RobotCar Dataset \cite{RadarRobotCarDatasetICRA2020} extends the original Oxford RobotCar Dataset \cite{maddern20171} with radar integration, captured during 32 urban traversals (280 km total) in January 2019 under diverse conditions (weather, traffic, lighting). Designed for robustness in sensor-challenging scenarios, it supports navigation and mapping research through multimodal sensor fusion. Recently, the Oxford Offroad Radar (OORD) dataset \cite{gadd2024oord} further complements \cite{RadarRobotCarDatasetICRA2020} with off-road radar data, filling the gap by offering data collected under extreme weather conditions in challenging mountainous terrains. This further benefits research that takes advantage of radar's operational advantages in difficult weather scenarios.

\subsection{Localization Against Public Data}

Several works have explored localizing ground sensors using overhead imagery, often referred to as \textit{crossview localization} \cite{Durgam_2024, Fervers_2023_CVPR}. Fervers et al. \cite{fervers2022continuous} achieved ~0.9 m accuracy on KITTI by projecting ground camera features into a BEV space using LiDAR and integrating the result into a Kalman filter with IMU. Zhu et al. \cite{Zhu2020AGCVLOAM} proposed AGCV-LOAM, using direct pose regression to register LiDAR with satellite imagery and jointly optimize radar odometry, achieving~4 m accuracy. Hu et al. \cite{Hu2024RoadStructure} generated road masks from LiDAR and extracted road lines from satellite images using gradient-based methods, achieving 2.5 m accuracy via template matching. Lee et al. \cite{Lee2024NoMap} combined learned place recognition and template matching for LiDAR-to-OpenStreetMap localization, using a particle filter and pose graph optimization, with accuracy ranging from 6~m to~0.2 m. Li et al. \cite{Li2023Geo} proposed a 2D-3D transformer for LiDAR-to-satellite registration, achieving 1.5 m accuracy. Ma et al. \cite{MaUSV2018} used handcrafted coastline features in 5 GHz marine radar and satellite images, aligning them via Hausdorff distance for~20 m accuracy. Bianchi et al. \cite{bianchi2021uav} used an autoencoder to compress satellite images and localized UAVs with stereo images, achieving 3 m RMSE. Fu et al. \cite{fu2020LiDAR} trained a model to extract features from LiDAR and satellite patches, then applied a particle filter to estimate vehicle pose.

\subsection{Radar-Satellite Localization}
 Several works have attempted to localize an imaging radar on the ground with overhead imagery. Tang et al. \cite{tang2020rslnet} propose an architecture, RSL-Net, capable of registering a single ground radar image with a proximal overhead satellite image. The network first predicts the rotation offset by taking a live radar scan, rotating it multiple times, and concatenating each rotated radar scan with a proximal satellite image. The following network stage generates a synthetic radar image using the rotation-aligned radar-satellite pair. Finally, the cross correlation between the generated image and the ground radar is computed to solve for the translation. Their follow-up work proposed a self-supervised version of this pipeline \cite{tang2021SelfSuper}, in which they also evaluated the feasibility of their pipeline as a standalone localization solution on an unseen test location. Their method contained errors of 40+ m with no quantitative evaluation. Tang et al. also proposed a method of metric localization and place recognition in \cite{Tang2023} for lidar and radar. Their method worked by learning an occupancy representation for the overhead imagery and obtaining a point cloud by raytracing from the center of the image. A learning-based point registration module is then used for metric localization, and the resulting embeddings are repurposed for place recognition. However, this system is limited to single-frame evaluation and does not demonstrate localization performance over an entire survey. There has also been work in localizing ground radar within OpenStreetMaps (OSM) \cite{Hong2023RadarOSM}, in which the authors use a radar odometry method and overhead registration using road segments, lane information, and surface building features provided by OSM in a filtering method. Part of their localization system includes restricting the vehicle's pose to lie on a road segment provided by OSM. This assumption in invalid in any off-road scenarios or a marine setting and brittle in areas that may not have been annotated with road segments or building information, as shown in Figure 6 in their work. The authors further optimized this work by incorporating semantic segmentation information from a LiDAR to improve radar odometry and registration by filtering the radar data \cite{Li2025Free}. There has also been work in registering satellite SAR radar data with satellite imagery to produce accurate terrain maps \cite{sommervold2023survey}, though this is not the focus of our work.

\section{METHODOLODY}
\begin{figure*}[t!]
    \centering
    \includegraphics[width=\textwidth]{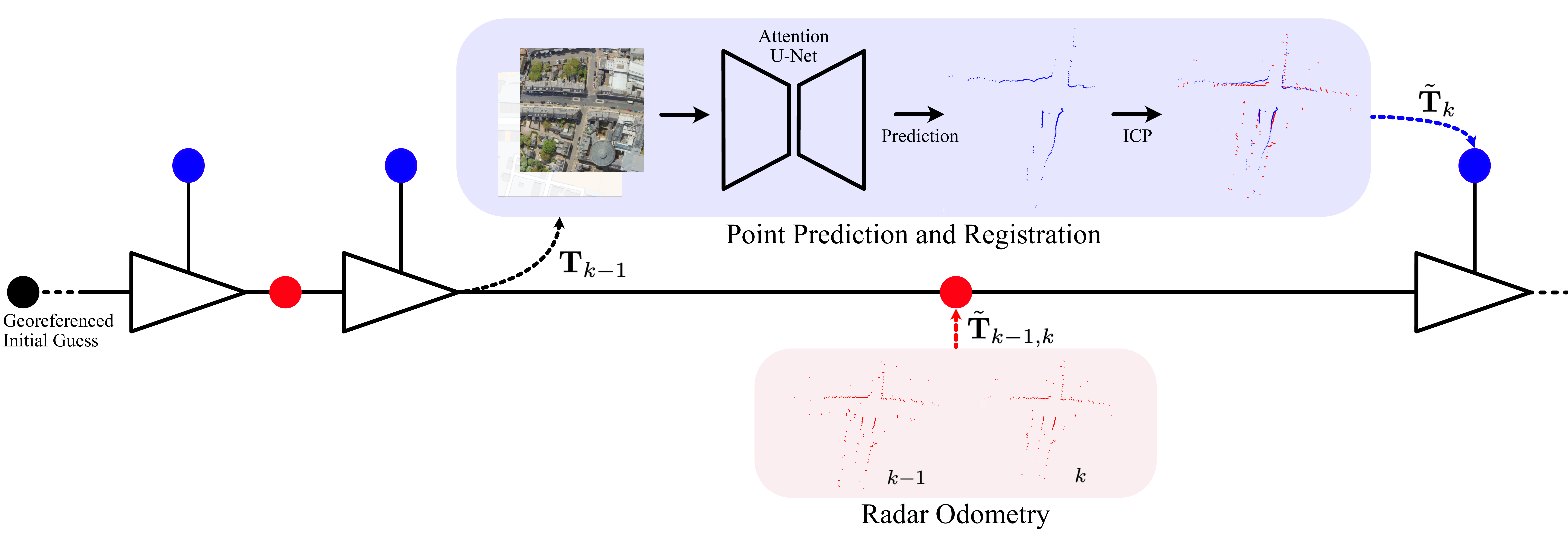} % Adjust filename and extension
    \caption{Overview of localization pipeline. The red factors are binary factors formed using the solution from point-to-point ICP between consecutive real radar scans. The blue factors are formed by using the most recent state estimate to retrieve a satellite image, indicated by $\Transform_{k-1}$, then extracting a predicted point cloud (blue) and computing a relative pose registration $\tilde{\Transform}_k$ from the live radar scan to the satellite image to form a global unary factor.}
% some how illustrate more of the window, make it more consistent with the code, show the blow ups at a single factor, add more to emphasize the window, maybe remove legend, explain why the dotted line is the way it is. Not obvious what the blue points are, no arrow between blue and purple boxes, no arrows on the lines
    \label{fig:overview}
\end{figure*}

Our method first consists of registering a single real radar scan with a proximal satellite image. First, we convert the two data sources to a common point cloud. To obtain a point cloud from the satellite image, we convert the satellite image into an occupancy image and then ray trace from the center of the image to obtain a point cloud. To obtain a point cloud from the radar image we employ the \textit{k-strongest} filtering method. We use ICP to compute the registration between these two point clouds. We then implement a localization solution that optimizes overhead registration with scan-to-scan radar odometry. An overview of this pipeline is illustrated in Figure \ref{fig:overview}.

\subsection{Learning To Predict Points} 

\begin{figure}[h]
    \centering
    \includegraphics[width=\linewidth]{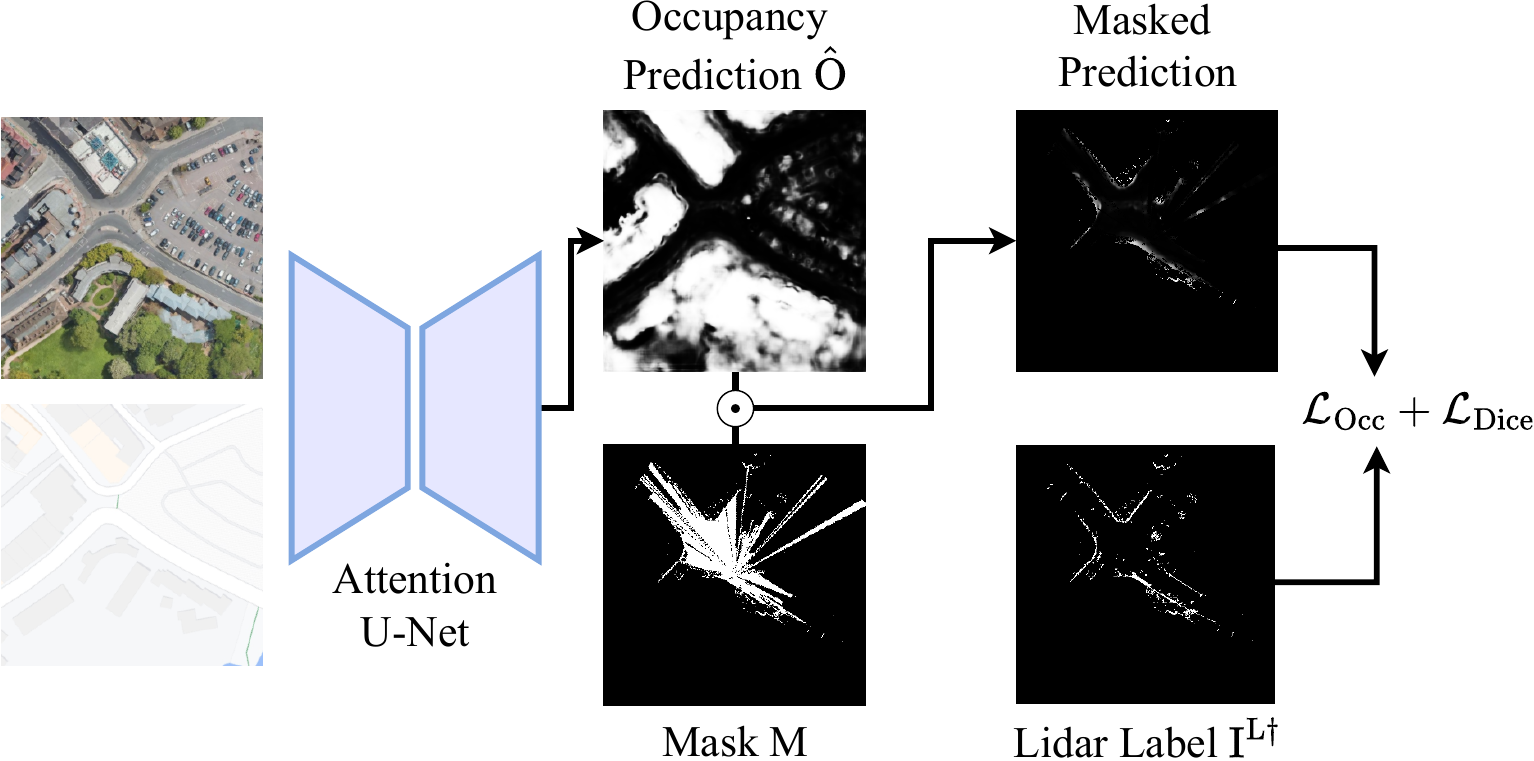}
    \caption{Overview of training pipeline for the learned occupancy representation where $\odot$ is an element-wise product. The network predictions are masked using $\text{M}$ to prevent gradients from flowing through predictions for which we are uncertain. The method for obtaining $\text{M}$ and $\text{I}^{L\dagger}$ is described in Section III-A. }
    \label{fig:occ_pipeline}
\end{figure}

This section summarizes the methodology in \cite{Tang2023}, in which the authors present a learning method to generate a point cloud given overhead imagery. To extract a point cloud from a satellite image, we first concatenate the RGB Overhead image and the Google Roadmaps image, which together we refer to as $\text{I}^{S}$ and pass it as input to an Attention U-Net \cite{oktay2018attentionunetlearninglook} to predict $\hat{\text{O}}$, a learned occupancy representation. We found the Attention U-Net to be better for extracting sharp edges along buildings and semantics such as dense foliage, as seen in Fig.~\ref{fig:occ_pipeline}. The predictions of this network are supervised using a binary lidar mask $\text{I}^{L\dagger}$ and a binary certainty mask $\text{M}$. Which leads to the following binary cross entropy loss term

\begin{equation} 
    \begin{aligned}[b]
 \mathcal{L}_{\text{Occ}} =& -\sum_{i,j} \text{M}(i,j) \big[ 
    \text{I}^{L\dagger}(i,j) \log \hat{\text{O}}(i,j) + \\
&    (1 - \text{I}^{L\dagger}(i,j)) \log (1 - \hat{\text{O}}(i,j)) 
\big].
\end{aligned}
\end{equation}

M is necessary since we are unsure of the occupancy in unseen parts of the satellite image. \noindent Due to the sparsity of labels provided by $\text{I}^{L\dagger}$, we also use a Dice loss to maximize the intersection over union. The medical segmentation community has heavily used Dice loss to segment small objects in situations with imbalanced annotations:

\begin{equation}
\mathcal{L}_{\text{Dice}}\hspace{-2pt} = \hspace{-2pt}1 \hspace{-1pt}- \hspace{-1pt}\frac{2 \sum_{i,j} \text{M}(i,j) \hat{\text{O}}(i,j) \text{I}^{L\dagger}(i,j)}{\sum_{i,j} \hspace{-1pt}\text{M}(i,j) \hat{\text{O}}(i,j) \hspace{-1pt}+\hspace{-1pt} \sum_{i,j}\hspace{-1pt} \text{M}(i,j) (\text{I}^{L\dagger}(i,j))}\hspace{-1pt}.
\end{equation}

We then optimize the occupancy network predictions $\hat{\text{O}}$ using a weighted combination of these losses,
\begin{equation}
\mathcal{L} = \mathcal{L}_{\text{Occ}} + \lambda \mathcal{L}_{\text{Dice}}.
\end{equation}

We found $\lambda = 0.5$ to yield the best results. To remove the ground points, we filter out all points below a height of 0 m in the sensor frame with $z$ pointing upwards. We also remove all points above a height of 3 m in the local sensor frame since the radar is a planar sensor with a narrow vertical beam width. $\text{M}$ is obtained by converting the lidar data into a bird's-eye view (BEV) polar representation, removing the ground points, tracing along each azimuth, and setting each pixel value to 1 until it is above a threshold $\tau_{\text{lidar}}$, which we set to 0.04. Finally, $\text{I}^{L\dagger}$ is obtained by removing ground points and binary intensity thresholding according to $\tau_{\text{lidar}}$. 

To obtain a point cloud at inference time, we transform $\hat{\text{O}}$ into a range azimuth polar representation and take the first point along each azimuth greater than $\tau_{\text{occ}}$, which we set to be 0.6. To obtain the polar representation, we keep the same $\frac{\rm m}{\rm pixel}$ as the cartesian image and assume 400 evenly spaced azimuths to match the radar sensor. For more details about this procedure, we refer the readers to \cite{Tang2023}. Although not evaluated in this work, this representation could allow for preprocessing the imagery over the entire survey area therefore requiring no GPU during deployment. Human annotators can manually adjust this representation in areas of incorrect predictions, which can further improve accuracy in safety-critical applications.

% \begin{figure}
%     \centering
%     % First row
%     \begin{subfigure}{\linewidth}  % Left figure in first row
%         \centering
%         \includegraphics[width=\linewidth]{figs/visualization_for_paper_0.png}
%     \end{subfigure}
%     %\hfill
%     \vspace{1mm} % Adjust vertical spacing if needed
%     \begin{subfigure}{\linewidth}  % Right figure in first row
%         \centering
%         \includegraphics[width=\linewidth]{figs/visualization_for_paper_1.png}
%     \end{subfigure}

%     % Second row
%     %\begin{subfigure}{0.48\linewidth}  % Left figure in second row
%     %    \centering
%     %    \includegraphics[width=\linewidth]{figs/trans_net_test_ex.png}
%     %    \caption{Regular U-Net}
%     %\end{subfigure}
%     %\hfill
%     %\begin{subfigure}{0.48\linewidth}  % Right figure in second row
%     %    \centering
%     %    \includegraphics[width=\linewidth]{figs/AttnUnet_ex.png}
%     %    \caption{Attention U-Net}
%     %\end{subfigure}

%     \caption{Comparison of U-Net vs Attention U-Net. The occupancy images on the left column are predic}
%     \label{fig:u-net comparison}
% \end{figure}
\subsection{Radar Point Extraction}
The authors in \cite{Tang2023} used a Cycle-GAN model trained on unpaired radar and lidar data to denoise radar data and extract similar geometry captured by both sensors. Doing this further increases runtime, adds more tunable parameters, and adds more potential for overfitting. Instead, we use the $k$-strongest filtering method to extract points for registration. $K$-strongest has been shown to be effective for mmWave radar odometry in extracting the salient geometry for localization and odometry \cite{prestonkrebs2024finerpointssystematiccomparison} \cite{adolfsson2022LiDAR}. $K$-strongest works by keeping the $k$ points with the highest intensity along each azimuth of the radar. In our experiments, we found that using a value of $k=9$ for radar to satellite registration, and $k=5$ for radar odometry, achieved the best results. 

\subsection{Radar to Satellite Registration}

Once the radar scan and the overhead image have been converted to a common modality, we compute the registration between the two point clouds using point-to-point ICP \cite{Besl92ICP}. The approach in \cite{Tang2023} uses a learning-based registration approach to register the two point clouds. However, using ICP offers more straightforward outlier rejection solutions and does not require a GPU during deployment. ICP has been shown to be effective for both localization and odometry problems \cite{BurnettAreWeReady2022} \cite{prestonkrebs2024finerpointssystematiccomparison} \cite{Adolfsson_2023} across both radar and lidar. ICP has also been shown to offer a more accurate solution with a close initial guess \cite{wang2019deepclosestpointlearning}, which is more suitable for precise localization. A common approach to outlier rejection is to include a robust cost or trim function in the ICP nearest-neighbors objective. Our experiments found that using a trim of 10 pixels (4.33 m) yields the best results. We also found in our experiments that initializing ICP with a larger trim distance (50 pixels) for five iterations helped recover from more significant errors while filtering out outlier registrations. Although the Google Roadmaps image is passed as an input to the occupancy model, we do not explicitly use any lane or building information as done in \cite{Hong2023RadarOSM} and \cite{Li2025Free}. This makes our method robust in areas where OSM lane information, such as a marine surface vessel, may be outdated or non-existent. Our learned representation can also recognize areas of dense vegetation that may appear in the ground radar, as shown in Figure \ref{fig:occ_pipeline}. 
%\sam{This is actually a very good strength maybe you can elaborate a little more...}

\subsection{Radar Odometry}
A key component of our localization is a form of radar odometry that can estimate smooth local motion. The odometry can also help localize the vehicle in challenging situations, such as overhead foliage or under short bridges. We use ICP between consecutive radar scans with a trim distance of 4 m for our radar odometry. We do not use robust cost functions, motion priors, motion compensation, key frame registration, or IMU measurements, as is common in other works. This is because our odometry does not need to be accurate over long distances for the vehicle to localize. Figure \ref{fig:traj_comparison} compares the estimates generated from only our odometry to estimates with our satellite correction factor. %For the odometry, we solve the following cost function for point-to-point ICP. \sam{need to include the cost...}

\subsection{Factor Graph Optimization}
To optimize our measurements from the odometry and overhead registration, we use the Batch Fixed-Lag Smoother in GTSam for online estimation \cite{gtsam}. We start our optimization with a georeferenced initial guess defined in WGS84 latitude and longitude with a heading in degrees from north, which we can define as an origin  $\CoordinateFrame{\rm map}$. We are optimizing the state variable, a $\LieGroupSE{2}$ metric transform at timestamp $k$, $\Transform_k$. This is the transformation from the vehicle frame $\CoordinateFrame{\rm v}$ to the map frame $\CoordinateFrame{\rm map}$, $\Transform_{\rm map,v}$. For odometry we use the solution from ICP between the filtered radar pointcloud at time $k-1$ and $k$ to estimate the transformation $\tilde{\Transform}_{k-1,k}$ and form the following binary cost factor:

\begin{equation}
    \textstyle J_{\text{odom},k}\hspace{-2pt} =\hspace{-2pt} \ln\hspace{-2pt}\left(\hspace{-1pt} \tilde{\mathbf{T}}_{k-1,k}  \mathbf{T}_{k-1,k}^{-1} \hspace{-2pt}\right)^{\vee^T}\hspace{-8pt} \mathbf{R}^{-1} \ln\hspace{-2pt}\left(\hspace{-1pt} \tilde{\mathbf{T}}_{k-1,k} \mathbf{T}_{k-1,k}^{-1} \hspace{-2pt}\right)^\vee\hspace{-5pt},
\end{equation}
where $\ln$ is the matrix logarithm map and $\vee$ is the inverse of the skew-symmetric operator, which turns a transform in $\LieGroupSE{2}$ to an $\Real{}^3$ vector in $\LieAlgebraSE{2}$. $\mathbf{\mathbf{R}}$ is an isotropic covariance matrix composed of $\sigma_{\text{odom}_{\text{xy}}}$ for $x,y$ noise and $\sigma_{\text{odom}_{\text{yaw}}}$ for yaw noise. For localization, we use the latest estimate $\Transform_{k-1}$ to fetch the satellite imagery defined in the map frame. We then predict the point cloud from this satellite image and use ICP to compute the registration between the live scan and the map to obtain $\tilde{\Transform}_{k}$ and form the following unary cost factor:

\begin{equation}
   J_{\text{loc},k} = \ln\left( \tilde{\mathbf{T}}_k  \mathbf{T}_k^{-1} \right)^{\vee^{T}} \mathbf{Q}^{-1} \ln\left( \tilde{\mathbf{T}}_k  \mathbf{T}_k^{-1} \right)^{\vee}.
\end{equation}
$\mathbf{\mathbf{Q}}$ is constructed in the same way as $\mathbf{R}$ with $\sigma_{\text{sat}}$ parameters. Some areas of the satellite imagery can produce poor registration due to overhanging trees in the roadway, construction, or overhead bridges. In our experiments, we found an effective way to mitigate this was to filter out measurements with a low match fitness score, which is defined as,
\begin{equation}
    \gamma_{\text{fit}} = \frac{N_{\text{inliers}}}{N_{\text{target}}},
\end{equation}
where $N_{\text{inliers}}$ is the number of matched inliers determined by ICP and $N_{\text{target}}$ is the number of points in the target. We use a threshold $\tau_{\text{fit}} = 0.6$ to determine whether or not the registration gets added to the factor graph.

We then jointly optimize these two factors over a sliding window of size $W$, which contains the last 10 seconds of estimated poses, to form our final cost we seek to minimize,

\begin{equation}
   J_k = \sum_{j=k-W}^{k}J_{\text{odom},j} + \sum_{j=k-W}^{k}J_{\text{loc},j}.
\end{equation}

\section{Experiments}

We train and evaluate our method on three datasets across a diverse set of geographic conditions, the Boreas Dataset \cite{burnett2023boreas}, the Oxford Radar Robot Car Dataset \cite{RadarRobotCarDatasetICRA2020}, and our own dataset collected on an Otter Unmanned Surface Vehicle (USV) in Northern Ontario, which we will refer to as the Boat dataset. The Boreas Dataset features repeated traversals over the same geographic location in the suburbs of Toronto. It has a 64-beam Velodyne lidar and Navtech RAS-6 sensor and post-processed RTX ground truth with an accuracy of $2$ cm. The Oxford dataset features repeated traversals of a 9 km route through the streets of Oxford and contains data from 2 64-beam Velodyne Sensors, a Navtech RAS-6 sensor, and NovAtel RTK GPS. The boat dataset contains lidar data from an Ouster OS1-128, a Navtech RAS-6 Radar, and post-processed Novatel GPS PwrPak7 for ground truth. All Navtech radars have a range resolution of 0.043 $\frac{\rm m}{\rm pixel}$, and have a maximum range of 140 m as we use satellite imagery with a resolution of 640x640 at 0.433 $\frac{\rm m}{\rm pixel}$ during evaluation.

\begin{figure}
    \includegraphics[width=\linewidth]{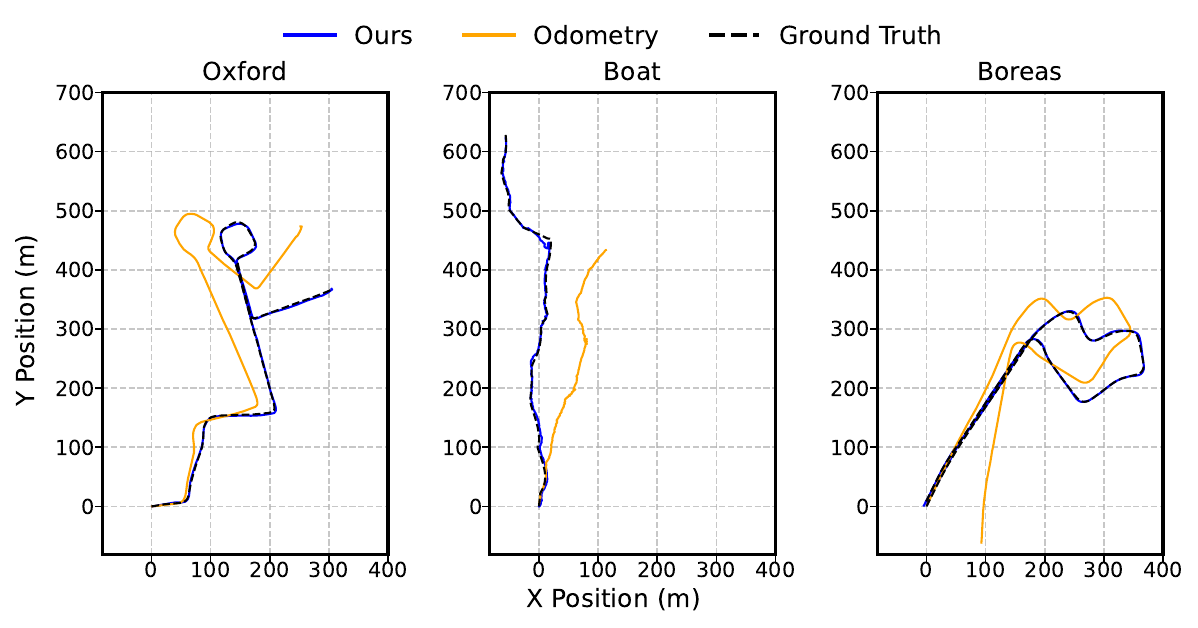}   
    \caption{Odometry drift visualization. In this comparison, we show our optimization without the satellite localization factor (orange) and with (blue). The figure shows our satellite localization factor significantly improves performance in unseen areas and does not require highly accurate odometry. The degraded boat odometry is discussed in Section IV-C.}
    \label{fig:traj_comparison}
\end{figure}

\begin{figure}
   \centering
   \begin{subfigure}{0.3\linewidth}
       \centering
       \includegraphics[width=\textwidth]{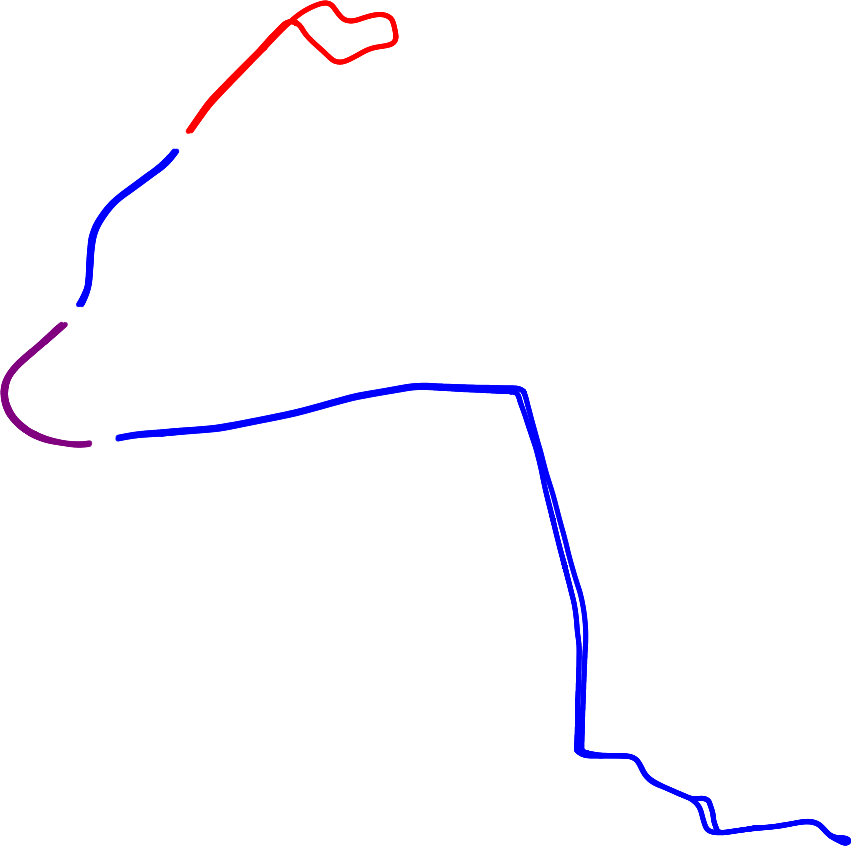}
       \caption{Boreas 8 km}
       \label{fig:fig1}
   \end{subfigure}
   %\hfill
   \begin{subfigure}{0.3\linewidth}
       \centering
       \includegraphics[width=\textwidth]{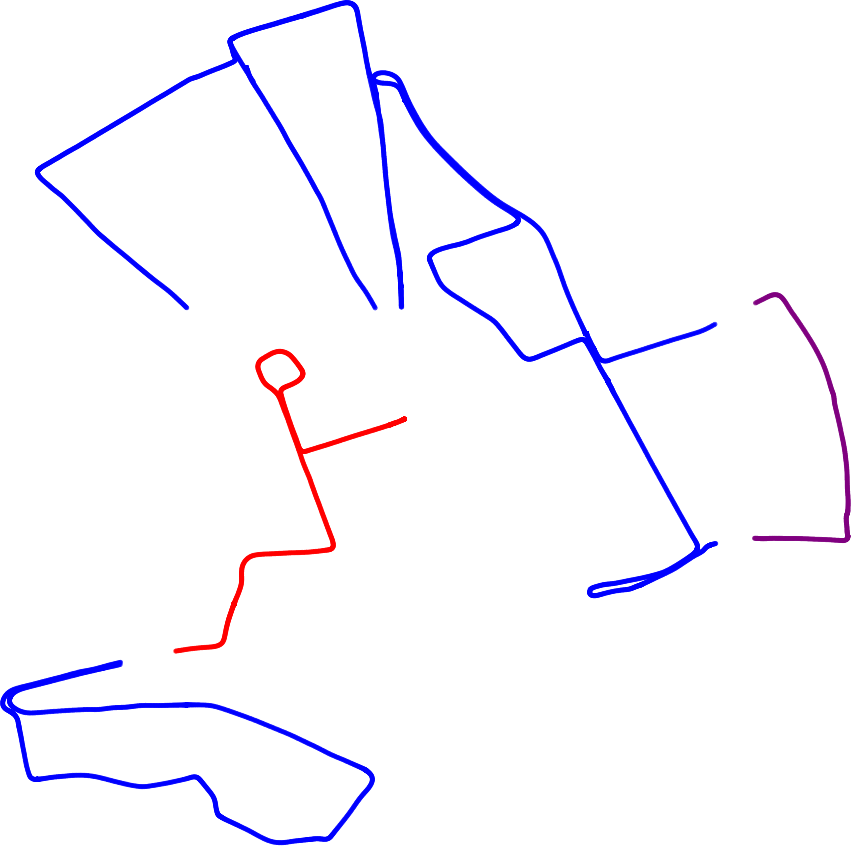}
       \caption{Oxford 10 km}
       \label{fig:fig2}
   \end{subfigure}
   \begin{subfigure}{0.3\linewidth}
       \centering
       \includegraphics[width=\textwidth]{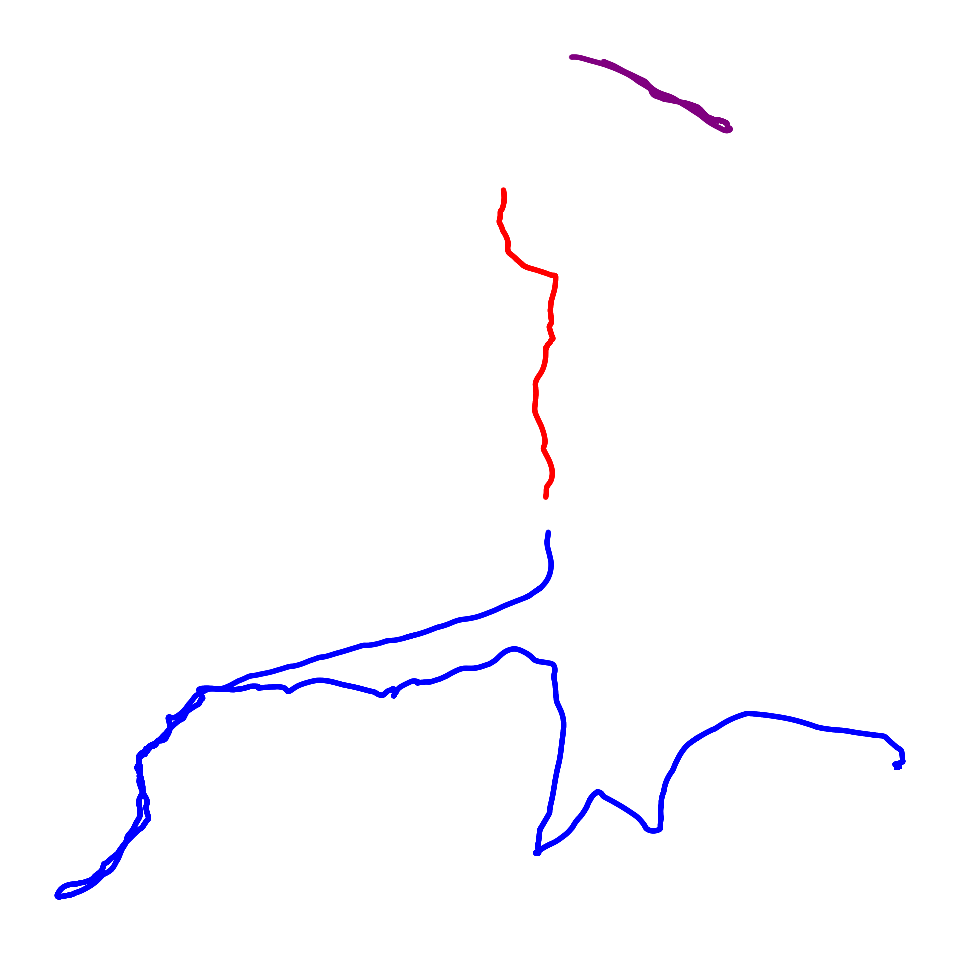}
       \caption{Boat 6 km}
       \label{fig:fig2}
   \end{subfigure}
   \caption{Dataset geographic splits. Blue is training, red is test, purple is the validation area. Path length is indicated for each dataset.}
    \label{fig:splits}
\end{figure}

\subsection{Training Details} 
For training the occupancy network we split up the datasets geographically according to Figure \ref{fig:splits} and aggregated the lidar data in multiple sequences when possible. Training and testing locations were chosen in Oxford according to \cite{tang2020rslnet} \cite{Tang2023} \cite{tang2021SelfSuper}. In Boreas, the test location was selected as a continuous piece of the trajectory so that we could evaluate our method from the start of the test sequence to the end without any geographical data leakage. However, for the boat data, there was a small portion of the sequence in which the boat crossed a region far from the shore between the current test and the validation set shown in Figure \ref{fig:splits}c, which is discarded as there were not sufficient features for localization. For the specific sequences, we used Boreas sequences 10-15-12-35 and 10-22-11-36, a single Boat sequence, and Oxford sequences 01-10-12-32-52, 01-10-14-50-05, and 01-10-15-19-41, which amounts to 106,576 lidar-satellite image pairs in the training set shown in blue in Figure \ref{fig:splits}. We use the Google Maps API to retrieve satellite and roadmap imagery for both locations. We use a resolution of $0.433 
\,\frac{\rm m}{\rm pixel}$, which roughly corresponds to a zoom level of 18 at a size of 320x320. We apply ColorJitter augmentations to the satellite images with a threshold of 0.1 for hue, saturation, and brightness. We train for 50 epochs using the Adam optimizer \cite{kingma2017adam}. We use the checkpoint that achieved the lowest translational ICP error in the Oxford validation location with the extracted radar points for our evaluations. 

\begin{figure}[h]
    \centering
    % First row
    \begin{subfigure}{\linewidth}  % Left figure in first row
        \centering
        \includegraphics[width=\linewidth]{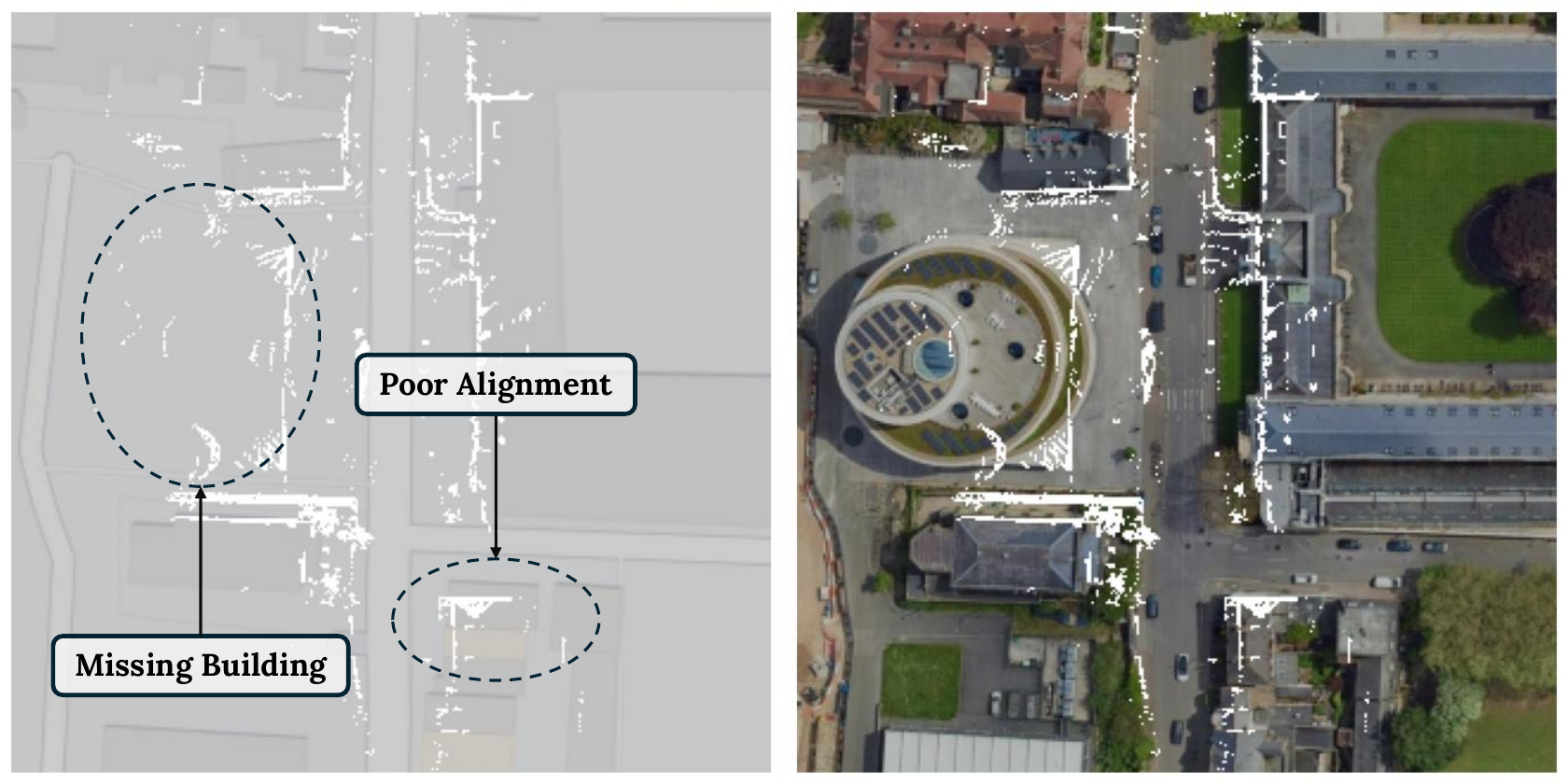}
    \end{subfigure}
    %\hfill
    %\vspace{1mm} % Adjust vertical spacing if needed
    \begin{subfigure}{\linewidth}  % Right figure in first row
        \centering
        \includegraphics[width=\linewidth]{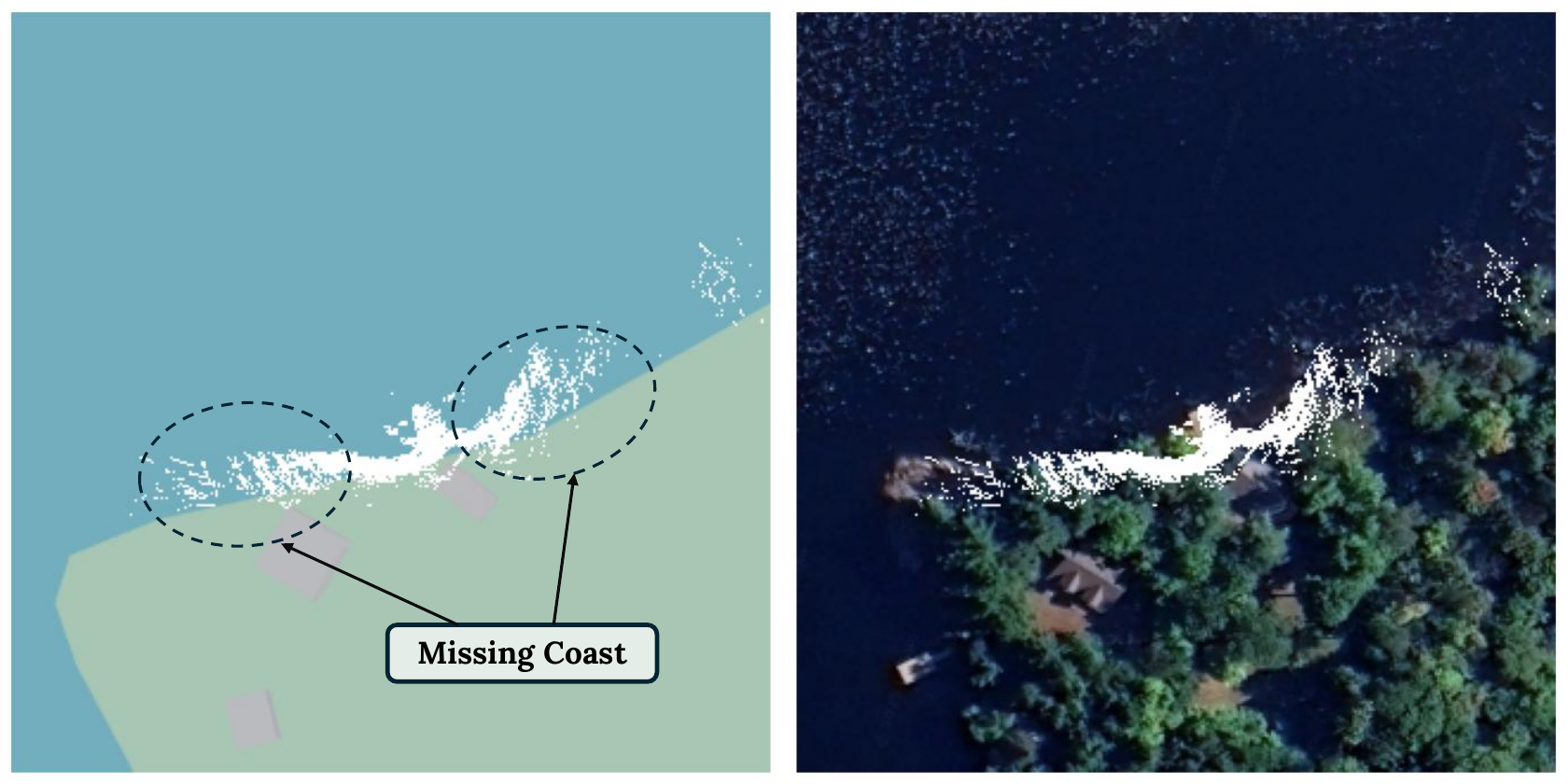}
    \end{subfigure}

    % Second row
    %\begin{subfigure}{0.48\linewidth}  % Left figure in second row
    %    \centering
    %    \includegraphics[width=\linewidth]{figs/trans_net_test_ex.png}
    %    \caption{Regular U-Net}
    %\end{subfigure}
    %\hfill
    %\begin{subfigure}{0.48\linewidth}  % Right figure in second row
    %    \centering
    %    \includegraphics[width=\linewidth]{figs/AttnUnet_ex.png}
    %    \caption{Attention U-Net}
    %\end{subfigure}

    \caption{Comparison of roadmap imagery with overhead RGB for alignment with ground lidar. The top row is taken from the Oxford Dataset. The bottom row shows the start of the Boat dataset. This illustrates how roadmap imagery is suboptimal for localization, especially in a marine scenario where the coastline is poorly represented. All images are fetched using Google Maps API \cite{GoogleMapsAPI}.}
    \label{fig:road map comparison}
\end{figure}

% \subsection{Single Frame Registration}

% To the best of the authors knowledge only \cite{tang2021SelfSuper} \cite{tang2020rslnet} \cite{Tang2023} provide results for localizing a single radar scan with a proximal overhead image. We mimic their evaluation by geographically partitioning the oxford dataset as shown in fig. as close as possible. We then retrieve a satellite image using the ground truth GPS coordinates of each radar scan, and for each radar scan we randomly add an offset of $\pm 25$ pixels along with a heading offset of $ \pm 22.5^{\circ}$.

\renewcommand{\arraystretch}{1.2}
\begin{table}[ht]
\centering
\begin{minipage}{\columnwidth}
\centering
\vspace{0.15cm}
\caption{Comparison across whole trajectory. To compare with previous work, we evaluate our method across the whole trajectory of each dataset. The * indicates that the occupancy predictor has been trained on a portion of the satellite images within the sequence, and our method also uses a history of frames in a window, whereas the other methods use filtering. The competing methods use road segments and can not be applied to the Boat data. The large difference in results across datasets is discussed in Section IV-C.}
\resizebox{\textwidth}{!}{%
\begin{tabular}{|c|cccc|cccc|c|}
\hline
\textbf{Method} & \multicolumn{4}{c|}{\textbf{Boreas}} & \multicolumn{4}{c|}{\textbf{Oxford}} & \textbf{Boat} \\
\cline{2-10}
                & 01 & 02 & 03 & 04 & 01 & 02 & 03 & 04 & 01 \\
\hline
\textbf{Hong \cite{Hong2023RadarOSM}} & 4.7 & 4.9 & 4.9 & 3.8 & 5.4 & 3.4 & 4.1 & 4.1 & -\\
\textbf{Li \cite{Li2025Free}}         & 3.0 & 3.0 & 3.4 & 3.7 & \textbf{4.5} & \textbf{3.3} & \textbf{3.3} & \textbf{3.5} & -\\
\textbf{Ours*}                        & \textbf{1.3} & \textbf{1.5} & \textbf{1.6} & \textbf{1.6} & \textbf{4.5} & 4.5 & 4.5 & \textbf{3.5} & \textbf{3.5} \\
\hline
\end{tabular}%
}
\label{tab:comparison}
\end{minipage}
\end{table}

\renewcommand{\arraystretch}{1.2}
\begin{table}[ht]
\caption{Test area evaluation. All values are RMSE with respect to ground truth in meters, in the unseen test area. Table~\ref{tab:comparison} computes the errors across the entire trajectory.} 
\centering
 \label{tab:comparison_test}

\begin{tabular}{|c|cccc|cccc|c|}
\hline
\textbf{Method} & \multicolumn{4}{c|}{\textbf{Boreas Sequences}} & \multicolumn{4}{c|}{\textbf{Oxford Sequences}} & \multicolumn{1}{c|} {\textbf{Boat}}\\
\cline{2-10}
                                      & 01 & 02 & 03 & 04 & 01 & 02 & 03 & 04 & 01 \\
\hline
\textbf{Ours}                         & 1.5 & 1.4 & 1.4 & 2.6 & 4.1 & 2.8  &  2.7  & 3.6 &  4.8  \\
\hline
\end{tabular}
%\caption*{}

\end{table}
\renewcommand{\arraystretch}{1.2}
\begin{table}[h]
\centering
\begin{minipage}{\columnwidth}  % Adjust 0.45 as desired
\captionsetup{width=\columnwidth}   % The caption will now wrap at the minipage width
\caption{Average RMSE for latitude, longitude, and heading in test locations (shown in red in Figure~\ref{fig:splits}), averaged across sequences in each dataset.}
\label{tab:avg_rmse}
\resizebox{\textwidth}{!}{%
\begin{tabular}{|c|c|c|c|}
\hline
\textbf{Dataset} & \textbf{Lat RMSE (m)} & \textbf{Long RMSE (m)} & \textbf{Yaw RMSE (°)} \\
\hline
Boreas & 1.08 & 1.30 & 3.13 \\
Oxford & 1.90 & 2.68 & 1.98 \\
Boat   & 2.27 & 4.18 & 4.44 \\
\hline
\end{tabular}%
}
\end{minipage}
\end{table}

% Add scale to the figures, maybe put length in subcaption
% Scale bar for satellite images

\subsection{Full Sequence Localization}

We compare our method directly with the closest works to ours: \cite{Hong2023RadarOSM} \cite{Li2025Free}. These methods compute the translational Root Mean Square Error (RMSE) in meters with respect to the GPS ground truth. The method used in \cite{tang2021SelfSuper} was tested as a localization solution on the Oxford dataset and errors of 40+ m on a single test set sequence with no quantitative evaluation (Section VI of the IJRR paper)\cite{tang2021SelfSuper}. Thus, we do not include it as a baseline due to a lack of comparable results. The authors in \cite{tang2020rslnet} and \cite{Tang2023} only conduct single-frame evaluations and do not evaluate their method as a standalone localization solution, and are also therefore not comparable. We use identical sequences in the \cite{Li2025Free} evaluation and conduct a comparison in Table \ref{tab:comparison}. Although our estimates are optimized in a sliding window fashion, we only take the latest estimate for evaluation to show the performance of our method in real-time and to keep the evaluation as fair as possible. Table \ref{tab:comparison_test} conducts the same evaluation on the test area designated in red in Figure \ref{fig:splits}. Table \ref{tab:avg_rmse} conducts a more detailed evaluation of our method in translation and heading. The results in this table contain only estimates in the testing area and are averaged across all sequences.

%\begin{figure}[ht]
%    \centering
%    \begin{subfigure}{0.32\linewidth}
%        \centering
%        \includegraphics[width=\textwidth]{figs/boreas_split.pdf}
%        \caption{Boreas}
%        \label{fig:fig1}
%    \end{subfigure}
%    \hfill
%    \begin{subfigure}{0.32\linewidth}
%        \centering
%        \includegraphics[width=\textwidth]{figs/Oxford_split.pdf}
%        \caption{Oxford}
%        \label{fig:fig2}
%    \end{subfigure}
%    \hfill
%    \begin{subfigure}{0.32\linewidth}
%        \centering
%        \includegraphics[width=\textwidth]{figs/boreas_split.pdf}
%        \caption{Moana}
%        \label{fig:fig3}
%    \end{subfigure}
%    \hfill
%    \caption{Dataset geographic splits}
%    \label{fig:three_figures}
%\end{figure}

\begin{figure*}[t]
    \centering

    \begin{minipage}[t]{0.33\textwidth}
        \centering
        \includegraphics[width=\textwidth]{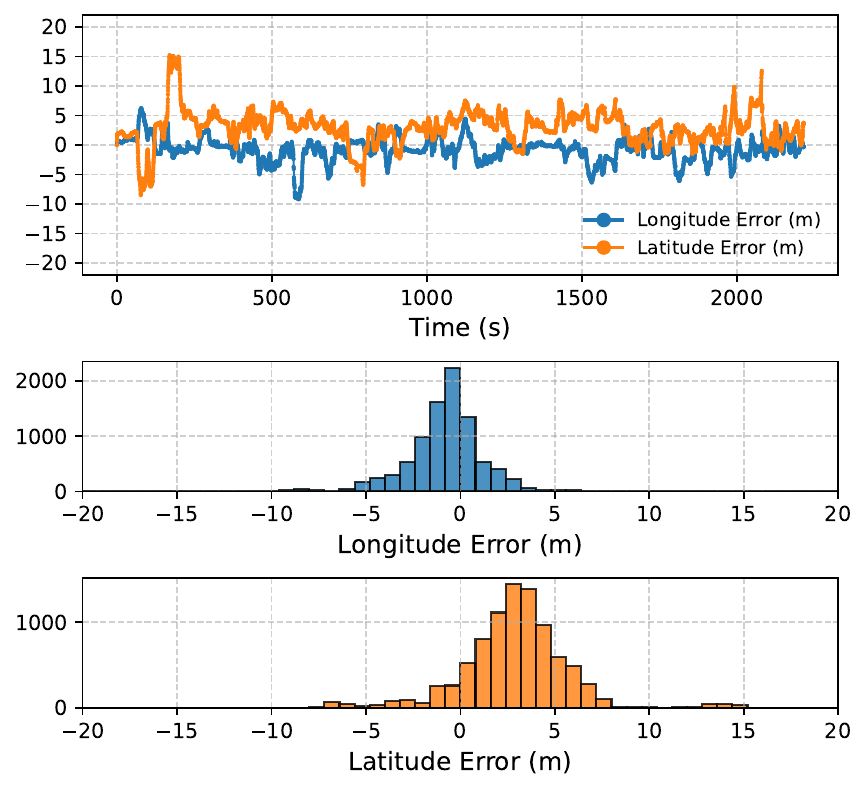}
        \caption*{(a) Oxford}
        \label{fig:oxford}
    \end{minipage}%
    \hfill
    \begin{minipage}[t]{0.33\textwidth}
        \centering
        \includegraphics[width=\textwidth]{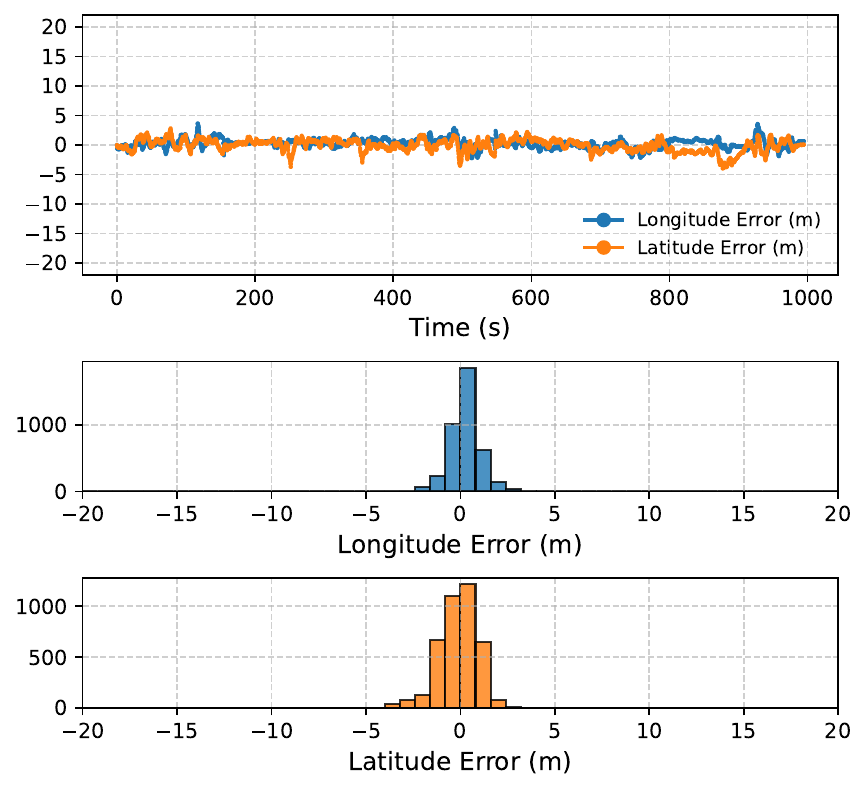}
        \caption*{(b) Boreas}
        \label{fig:boreas}
    \end{minipage}%
    \hfill
    \begin{minipage}[t]{0.33\textwidth}
        \centering
        \includegraphics[width=\textwidth]{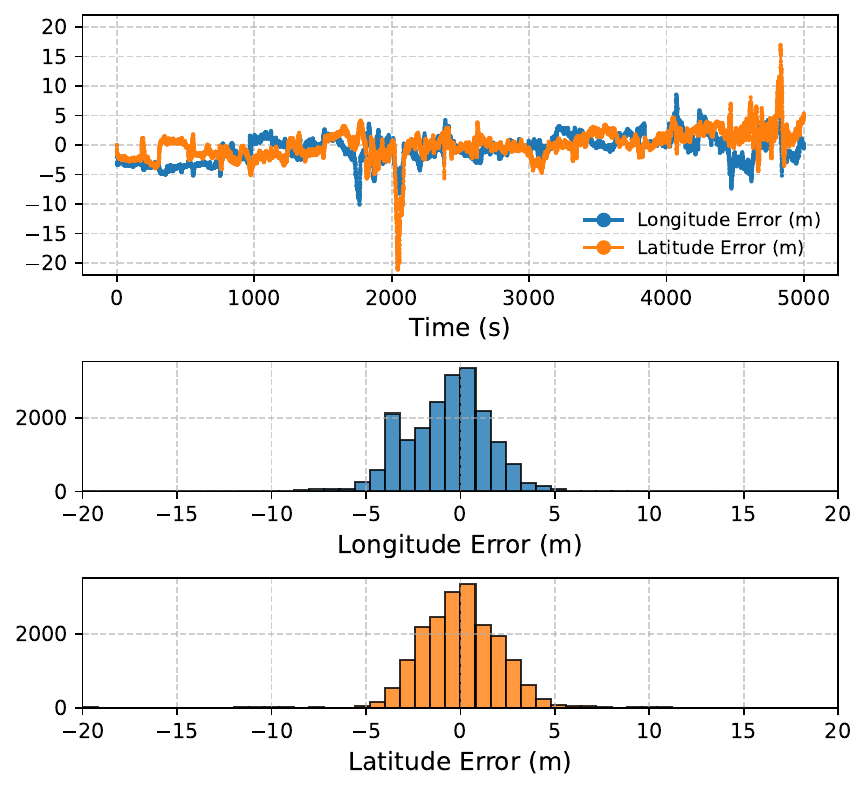}
        \caption*{(c) Boat}
        \label{fig:oxford_dup}
    \end{minipage}

    \caption{Localization error analysis: The top of each subfigure contains the latitude and longitude errors in meters vs time in seconds. As the error plots show, our method can recover from large errors caused by occlusions and poor overhead imagery. Under each error plot is a histogram of the latitude and longitude errors, which contain nearly Gaussian errors and with bias in some locations.}
    \label{fig:error_stack}
\end{figure*}

% \vspace{-1.5cm}
\subsection{Discussion of Results and Lessons Learned}
Our evaluation shows that our method outperforms previous radar-to-overhead imagery methods on Boreas while remaining competitive on the challenging Oxford dataset. Our method can also recover when errors accumulate from degenerate overhead imagery and heavy occlusion. One reason for the significant performance gap between the Oxford and Boreas data could be the large occlusions caused by moving vehicles, which are much closer to the ego vehicle in Oxford. Oxford is also an urban area with taller buildings, leading to more significant orthorectification artifacts that can destabilize training and cause larger errors. The Boreas data also contains post-processed GPS ground truth, leading to more stable occupancy model training. For the Boat data, the radar range was too short to capture $360^{\circ}$ of the shore, and large portions of the sequence only captured less than $180^{\circ}$ of the shore, leading to suboptimal constraints on the localization. However, our method works out of the box in a marine environment containing no road segments requiring minimal tuning of the $\sigma_{\text{odom}}$ covariances and increasing $k_{\text{odom}}$ and $k_{\text{sat}}$ to 20 due to the radar capturing lots of erroneous points from various leaves and foliage. The degraded odometry performance shown in Figure \ref{fig:traj_comparison} on the marine data could be due to a lack of coast features and slow movement compared to the road data. Our data collection platform also experienced packet loss, leading to many spurious detections.
%\sam{how competitive like 20 vs 21? need numbers}

\begin{table}[h]
\caption{Key Localization Parameters Used in the System}
\label{tab:params}
\centering
\begin{tabular}{lll}
\hline
\textbf{Parameter} & \textbf{Description} & \textbf{Value} \\
\hline
$\sigma_{\text{odom}_{\text{xy}}}$ & Standard deviation of odometry noise & 0.04 m \\
$\sigma_{\text{odom}_{\text{yaw}}}$ & Standard deviation of odometry noise & $0.1^\circ$ \\
$\sigma_{\text{sat}_{\text{xy}}}$ & Standard deviation of overhead noise & 0.5 m \\
$\sigma_{\text{sat}_{\text{yaw}}}$ & Standard deviation of overhead noise & $4.5^{\circ}$ \\
$\delta_{\text{sat}}$ & Trim distance for satellite ICP & 4.33 m \\
$\delta_{\text{radar}}$ & Trim distance for radar ICP  & 4 m\\
$\tau_{\text{fit}} $ & Threshold for ICP fitness score & 0.6 \\
$k_{\text{radar}}$  & K Strongest radar odometry & 5 \\
$k_{\text{sat}}$  & K Strongest radar overhead & 9 \\
%plz fix $k_{\rm sat}$  & K Strongest Radar to Sat & 9 \\
$\tau_{\text{occ}}$ & Occupancy Threshold & 0.6 \\

\hline
\end{tabular}
\end{table}

% \begin{figure*}[t]
%     \centering

%     \begin{minipage}[t]{0.32\textwidth}
%         \centering
%         \includegraphics[width=\textwidth]{figs/err_analysis_oxford.pdf}
%         \caption*{(a) Oxford}
%         \label{fig:oxford}
%     \end{minipage}%
%     \hfill
%     \begin{minipage}[t]{0.32\textwidth}
%         \centering
%         \includegraphics[width=\textwidth]{figs/err_analysis_boreas.pdf}
%         \caption*{(b) Boreas}
%         \label{fig:boreas}
%     \end{minipage}%
%     \hfill
%     \begin{minipage}[t]{0.32\textwidth}
%         \centering
%         \includegraphics[width=\textwidth]{figs/err_analysis_boat.pdf}
%         \caption*{(c) Boat}
%         \label{fig:oxford_dup}
%     \end{minipage}

%     \caption{Localization error analysis: The top of each subfigure contains the X and Y errors vs the frame ID. As the error plots show, our method can recover from large errors caused by occlusions and poor overhead imagery. Under each error plot is a histogram of the X and Y errors, which contain nearly Gaussian errors, exhibiting bias in some locations.}
%     \label{fig:error_stack}
% \end{figure*}

\section{CONCLUSIONS AND FUTURE WORK}
In this work, we presented a learning-based localization solution for mobile platforms that jointly optimizes radar odometry and overhead RGB satellite image registration. Our method enables GNSS-free global localization in previously unseen areas using only radar and publicly available satellite imagery. The system has several limitations that point to future work. It occasionally struggles with zero-velocity estimation when the platform is stationary, especially in the presence of dynamic occluders such as passing vehicles. Incorporating wheel encoder data could help improve robustness in these scenarios. Localization performance may also benefit from longer-range radar, particularly in marine environments where the vessel navigates far from shore and radar returns are sparse. Occlusion challenges could be mitigated by computing registration with small submaps, allowing for the removal of dynamic objects.
Additionally, training on public self-driving datasets that provide lidar and GPS ground truth could improve the performance of the radar occupancy network. A promising extension involves fine-tuning the occupancy network using a differentiable point extraction method, and integrating a form of differentiable ICP could enhance radar-to-satellite registration accuracy. Another potentially valuable extension to this work could be a systematic comparison of different satellite imagery providers to analyze which source of imagery offers the best ground localization performance. Previous such studies have been conducted for OpenStreetMaps and have quantified significant variance in the accuracy of these maps across geographic locations \cite{ijgi7080289}. Overall, our system provides a step toward reliable, GPS-free localization for mobile platforms operating in unstructured and unmapped environments.

\section*{ACKNOWLEDGMENT}
This work was supported by NSERC, Navtech, and Hexagon NovAtel. 

%\newpage
\IEEEtriggeratref{0}
\bibliographystyle{./IEEEtranBST/IEEEtran}
\bibliography{./IEEEtranBST/IEEEabrv,./references}

\begin{thebibliography}{10}
\providecommand{\url}[1]{#1}
\csname url@rmstyle\endcsname
\providecommand{\newblock}{\relax}
\providecommand{\bibinfo}[2]{#2}
\providecommand\BIBentrySTDinterwordspacing{\spaceskip=0pt\relax}
\providecommand\BIBentryALTinterwordstretchfactor{4}
\providecommand\BIBentryALTinterwordspacing{\spaceskip=\fontdimen2\font plus
\BIBentryALTinterwordstretchfactor\fontdimen3\font minus
  \fontdimen4\font\relax}
\providecommand\BIBforeignlanguage[2]{{%
\expandafter\ifx\csname l@#1\endcsname\relax
\typeout{** WARNING: IEEEtran.bst: No hyphenation pattern has been}%
\typeout{** loaded for the language `#1'. Using the pattern for}%
\typeout{** the default language instead.}%
\else
\language=\csname l@#1\endcsname
\fi
#2}}

\bibitem{BurnettAreWeReady2022}
K.~Burnett, Y.~Wu, D.~J. Yoon, A.~P. Schoellig, and T.~D. Barfoot, ``Are we
  ready for radar to replace lidar in all-weather mapping and localization?''
  \emph{IEEE Robotics and Automation Letters}, vol.~7, no.~4, pp.
  10\,328--10\,335, 2022.

\bibitem{skolnik1962introduction}
M.~I. Skolnik, ``Introduction to radar,'' \emph{Radar handbook}, vol.~2, p.~21,
  1962.

\bibitem{s21165397}
J.~Vargas, S.~Alsweiss, O.~Toker, R.~Razdan, and J.~Santos, ``An overview of
  autonomous vehicles sensors and their vulnerability to weather conditions,''
  \emph{Sensors}, vol.~21, no.~16, 2021.

\bibitem{qiao2024radar}
X.~Qiao, A.~Krawciw, S.~Lilge, and T.~D. Barfoot, ``Radar teach and repeat:
  Architecture and initial field testing,'' \emph{arXiv preprint
  arXiv:2409.10491}, 2024.

\bibitem{Fervers_2023_CVPR}
F.~Fervers, S.~Bullinger, C.~Bodensteiner, M.~Arens, and R.~Stiefelhagen,
  ``Uncertainty-aware vision-based metric cross-view geolocalization,'' in
  \emph{Proc. IEEE/CVF Conf. Comput. Vision Pattern Recognit. (CVPR)}, June
  2023, pp. 21\,621--21\,631.

\bibitem{Tang2023}
T.~Y. Tang, D.~De~Martini, and P.~Newman, ``Point-based metric and topological
  localisation between lidar and overhead imagery,'' \emph{Autonomous Robots},
  vol.~47, no.~5, pp. 595--615, Jun 2023.

\bibitem{fervers2022continuous}
F.~Fervers, S.~Bullinger, C.~Bodensteiner, M.~Arens, and R.~Stiefelhagen,
  ``Continuous self-localization on aerial images using visual and lidar
  sensors,'' in \emph{Proc. IEEE/RSJ Int. Conf. Intell. Robots Syst.
  (IROS)}.\hskip 1em plus 0.5em minus 0.4em\relax IEEE, 2022, pp. 7028--7035.

\bibitem{mcconnell2022overheadimagefactorsunderwater}
J.~McConnell, F.~Chen, and B.~Englot, ``Overhead image factors for underwater
  sonar-based slam,'' 2022.

\bibitem{tang2020rslnet}
T.~Y. Tang, D.~D. Martini, D.~Barnes, and P.~Newman, ``Rsl-net: Localising in
  satellite images from a radar on the ground,'' 2020.

\bibitem{tang2021SelfSuper}
T.~Y. Tang, D.~D. Martini, S.~Wu, and P.~Newman, ``Self-supervised learning for
  using overhead imagery as maps in outdoor range sensor localization,''
  \emph{The International Journal of Robotics Research}, vol.~40, no. 12-14,
  pp. 1488--1509, 2021, pMID: 34992328.

\bibitem{Hong2023RadarOSM}
Z.~Hong, Y.~Petillot, K.~Zhang, S.~Xu, and S.~Wang, ``Large-scale radar
  localization using online public maps,'' in \emph{2023 IEEE International
  Conference on Robotics and Automation (ICRA)}, 2023, pp. 3990--3996.

\bibitem{doviak2014doppler}
R.~J. Doviak and D.~S. Zrnic, \emph{Doppler radar \& weather
  observations}.\hskip 1em plus 0.5em minus 0.4em\relax Academic press, 2014.

\bibitem{fitch2012synthetic}
J.~P. Fitch, \emph{Synthetic aperture radar}.\hskip 1em plus 0.5em minus
  0.4em\relax Springer Science \& Business Media, 2012.

\bibitem{jol2008ground}
H.~M. Jol, \emph{Ground penetrating radar theory and applications}.\hskip 1em
  plus 0.5em minus 0.4em\relax elsevier, 2008.

\bibitem{4250281}
E.~Brookner, ``Phased-array and radar breakthroughs,'' in \emph{2007 IEEE Radar
  Conference}, 2007, pp. 37--42.

\bibitem{9513303}
R.~Vehmas and N.~Neuberger, ``Inverse synthetic aperture radar imaging: A
  historical perspective and state-of-the-art survey,'' \emph{IEEE Access},
  vol.~9, pp. 113\,917--113\,943, 2021.

\bibitem{Jang_2024}
H.~Jang, M.~Jung, M.-H. Jeon, and A.~Kim, ``Lodestar: Maritime radar descriptor
  for semi-direct radar odometry,'' \emph{IEEE Robotics and Automation
  Letters}, vol.~9, no.~2, p. 1684–1691, Feb. 2024.

\bibitem{Besl92ICP}
P.~Besl and N.~D. McKay, ``A method for registration of 3-d shapes,''
  \emph{IEEE Transactions on Pattern Analysis and Machine Intelligence},
  vol.~14, no.~2, pp. 239--256, 1992.

\bibitem{Cen2019Ego}
S.~H. Cen and P.~Newman, ``Radar-only ego-motion estimation in difficult
  settings via graph matching,'' in \emph{2019 International Conference on
  Robotics and Automation (ICRA)}, 2019, pp. 298--304.

\bibitem{adolfsson2022LiDAR}
D.~Adolfsson, M.~Magnusson, A.~Alhashimi, A.~J. Lilienthal, and H.~Andreasson,
  ``Lidar-level localization with radar? the cfear approach to accurate, fast,
  and robust large-scale radar odometry in diverse environments,'' \emph{IEEE
  Transactions on robotics}, vol.~39, no.~2, pp. 1476--1495, 2022.

\bibitem{barnes2020radarlearningpredictrobust}
\BIBentryALTinterwordspacing
D.~Barnes and I.~Posner, ``Under the radar: Learning to predict robust
  keypoints for odometry estimation and metric localisation in radar,'' 2020.
  [Online]. Available: \url{https://arxiv.org/abs/2001.10789}
\BIBentrySTDinterwordspacing

\bibitem{burnett2021radarodometrycombiningprobabilistic}
\BIBentryALTinterwordspacing
K.~Burnett, D.~J. Yoon, A.~P. Schoellig, and T.~D. Barfoot, ``Radar odometry
  combining probabilistic estimation and unsupervised feature learning,'' 2021.
  [Online]. Available: \url{https://arxiv.org/abs/2105.14152}
\BIBentrySTDinterwordspacing

\bibitem{barnes2020maskingmovinglearningdistractionfree}
\BIBentryALTinterwordspacing
D.~Barnes, R.~Weston, and I.~Posner, ``Masking by moving: Learning
  distraction-free radar odometry from pose information,'' 2020. [Online].
  Available: \url{https://arxiv.org/abs/1909.03752}
\BIBentrySTDinterwordspacing

\bibitem{schiller2022}
C.~H. Schiller, B.~Arsenali, D.~Maas, and S.~Maranó, ``Improving marine radar
  odometry by modeling radar resolution and exploiting additional temporal
  information,'' in \emph{2022 IEEE/RSJ International Conference on Intelligent
  Robots and Systems (IROS)}, 2022, pp. 8436--8441.

\bibitem{lisus2024doppler}
D.~Lisus, K.~Burnett, D.~J. Yoon, R.~Poulton, J.~Marshall, and T.~D. Barfoot,
  ``Are doppler velocity measurements useful for spinning radar odometry?''
  \emph{IEEE Robotics and Automation Letters}, 2024.

\bibitem{hong2021radar}
Z.~Hong, Y.~Petillot, A.~Wallace, and S.~Wang, ``Radar slam: A robust slam
  system for all weather conditions,'' \emph{arXiv preprint arXiv:2104.05347},
  2021.

\bibitem{sie2024radarize}
E.~Sie, X.~Wu, H.~Guo, and D.~Vasisht, ``Radarize: Enhancing radar slam with
  generalizable doppler-based odometry,'' in \emph{Proceedings of the 22nd
  Annual International Conference on Mobile Systems, Applications and
  Services}, 2024, pp. 331--344.

\bibitem{barnes2018driven}
D.~Barnes, W.~Maddern, G.~Pascoe, and I.~Posner, ``Driven to distraction:
  Self-supervised distractor learning for robust monocular visual odometry in
  urban environments,'' in \emph{2018 IEEE International Conference on Robotics
  and Automation (ICRA)}.\hskip 1em plus 0.5em minus 0.4em\relax IEEE, 2018,
  pp. 1894--1900.

\bibitem{Durgam_2024}
A.~Durgam, S.~Paheding, V.~Dhiman, and V.~Devabhaktuni, ``Cross-view
  geo-localization: A survey,'' \emph{IEEE Access}, vol.~12, p.
  192028–192050, 2024.

\bibitem{Zhu2020AGCVLOAM}
M.~Zhu, Y.~Yang, W.~Song, M.~Wang, and M.~Fu, ``Agcv-loam: Air-ground
  cross-view based lidar odometry and mapping,'' in \emph{2020 Chinese Control
  And Decision Conference (CCDC)}, 2020, pp. 5261--5266.

\bibitem{Hu2024RoadStructure}
D.~Hu, X.~Yuan, H.~Xi, J.~Li, Z.~Song, F.~Xiong, K.~Zhang, and C.~Zhao, ``Road
  structure inspired ugv-satellite cross-view geo-localization,'' \emph{IEEE
  Journal of Selected Topics in Applied Earth Observations and Remote Sensing},
  vol.~17, pp. 16\,767--16\,786, 2024.

\bibitem{Lee2024NoMap}
S.~Lee and J.-H. Ryu, ``Autonomous vehicle localization without prior
  high-definition map,'' \emph{IEEE Transactions on Robotics}, vol.~40, pp.
  2888--2906, 2024.

\bibitem{Li2023Geo}
L.~Li, Y.~Ma, K.~Tang, X.~Zhao, C.~Chen, J.~Huang, J.~Mei, and Y.~Liu,
  ``Geo-localization with transformer-based 2d-3d match network,'' \emph{IEEE
  Robotics and Automation Letters}, vol.~8, no.~8, pp. 4855--4862, 2023.

\bibitem{MaUSV2018}
H.~Ma, E.~Smart, A.~Ahmed, and D.~Brown, ``Radar image-based positioning for
  usv under gps denial environment,'' \emph{IEEE Transactions on Intelligent
  Transportation Systems}, vol.~19, no.~1, pp. 72--80, 2018.

\bibitem{bianchi2021uav}
M.~Bianchi and T.~D. Barfoot, ``Uav localization using autoencoded satellite
  images,'' \emph{IEEE Robotics and Automation Letters}, vol.~6, no.~2, pp.
  1761--1768, 2021.

\bibitem{fu2020LiDAR}
M.~Fu, M.~Zhu, Y.~Yang, W.~Song, and M.~Wang, ``Lidar-based vehicle
  localization on the satellite image via a neural network,'' \emph{Robotics
  and Autonomous Systems}, vol. 129, p. 103519, 2020.

\bibitem{Li2025Free}
S.~Li, Z.~Hong, Y.~Chen, L.~Hu, and J.~Qin, ``Get it for free: Radar
  segmentation without expert labels and its application in odometry and
  localization,'' \emph{IEEE Robotics and Automation Letters}, vol.~10, no.~3,
  pp. 2678--2685, 2025.

\bibitem{sommervold2023survey}
O.~Sommervold, M.~Gazzea, and R.~Arghandeh, ``A survey on sar and optical
  satellite image registration,'' \emph{Remote Sensing}, vol.~15, no.~3, p.
  850, 2023.

\bibitem{oktay2018attentionunetlearninglook}
\BIBentryALTinterwordspacing
O.~Oktay, J.~Schlemper, L.~L. Folgoc, M.~Lee, M.~Heinrich, K.~Misawa, K.~Mori,
  S.~McDonagh, N.~Y. Hammerla, B.~Kainz, B.~Glocker, and D.~Rueckert,
  ``Attention u-net: Learning where to look for the pancreas,'' 2018. [Online].
  Available: \url{https://arxiv.org/abs/1804.03999}
\BIBentrySTDinterwordspacing

\bibitem{prestonkrebs2024finerpointssystematiccomparison}
\BIBentryALTinterwordspacing
E.~Preston-Krebs, D.~Lisus, and T.~D. Barfoot, ``The finer points: A systematic
  comparison of point-cloud extractors for radar odometry,'' 2024. [Online].
  Available: \url{https://arxiv.org/abs/2409.12256}
\BIBentrySTDinterwordspacing

\bibitem{Adolfsson_2023}
D.~Adolfsson, M.~Magnusson, A.~Alhashimi, A.~J. Lilienthal, and H.~Andreasson,
  ``Lidar-level localization with radar? the cfear approach to accurate, fast,
  and robust large-scale radar odometry in diverse environments,'' \emph{IEEE
  Transactions on Robotics}, vol.~39, no.~2, p. 1476–1495, Apr. 2023.

\bibitem{wang2019deepclosestpointlearning}
Y.~Wang and J.~M. Solomon, ``Deep closest point: Learning representations for
  point cloud registration,'' 2019.

\bibitem{gtsam}
\BIBentryALTinterwordspacing
F.~Dellaert and G.~Contributors, ``borglab/gtsam,'' May 2022. [Online].
  Available: \url{https://github.com/borglab/gtsam)}
\BIBentrySTDinterwordspacing

\bibitem{burnett2023boreas}
K.~Burnett, D.~J. Yoon, Y.~Wu, A.~Z. Li, H.~Zhang, S.~Lu, J.~Qian, W.-K. Tseng,
  A.~Lambert, K.~Y.~K. Leung, A.~P. Schoellig, and T.~D. Barfoot, ``Boreas: A
  multi-season autonomous driving dataset,'' 2023.

\bibitem{RadarRobotCarDatasetICRA2020}
D.~Barnes, M.~Gadd, P.~Murcutt, P.~Newman, and I.~Posner, ``The oxford radar
  robotcar dataset: A radar extension to the oxford robotcar dataset,'' in
  \emph{Proc. IEEE Int. Conf. Robot. Autom. (ICRA)}, Paris, 2020.

\bibitem{kingma2017adam}
D.~P. Kingma and J.~Ba, ``Adam: A method for stochastic optimization,'' 2017.

\bibitem{GoogleMapsAPI}
\BIBentryALTinterwordspacing
{Google}, \emph{Google Maps JavaScript API v3 Reference}, Google, 2025.
  [Online]. Available:
  \url{https://developers.google.com/maps/documentation/javascript/reference}
\BIBentrySTDinterwordspacing

\bibitem{ijgi7080289}
M.~A. Brovelli and G.~Zamboni, ``A new method for the assessment of spatial
  accuracy and completeness of openstreetmap building footprints,'' \emph{ISPRS
  International Journal of Geo-Information}, vol.~7, no.~8, 2018.

\end{thebibliography}

\end{document}